\documentclass[conference,a4paper]{APSIPA2026}

\usepackage[table]{xcolor}            % enables \cellcolor in tabular
\newcommand{\best}[1]{\cellcolor{red!60}#1}
\newcommand{\second}[1]{\cellcolor{orange!57}#1}
\newcommand{\third}[1]{\cellcolor{yellow!55}#1}

\usepackage{amsmath,amssymb}
\usepackage{graphicx}
\usepackage{multirow}
\usepackage{threeparttable}
\usepackage{booktabs}
\usepackage{xspace}
\usepackage{xcolor}
\usepackage[backend=biber,style=ieee]{biblatex}
\usepackage{geometry}
\usepackage{fancyhdr}

\fancypagestyle{firststyle}{
  \fancyhf{}
  \fancyhead[C]{2026 Asia Pacific Signal and Information Processing Association Annual Summit and Conference (APSIPA ASC)}
}

\newcommand{\dgs}{D-3DGS\xspace}
\newcommand{\dnerf}{D-NeRF\xspace}
\newcommand{\nerfds}{NeRF-DS\xspace}

\newcommand{\cfc}{CfC\xspace}

\begin{document}

\title{Liquid Neural Networks as a Drop-in Continuous-Time Deformation Field for Dynamic 3D Gaussian Splatting}

\author{
\authorblockN{
Mingzhao Li\authorrefmark{1},\quad
Arghya Pal\authorrefmark{1},\quad
Guan Yuan Tan\authorrefmark{1}
}

\authorblockA{
\authorrefmark{1}
School of Information Technology, Monash University, Selangor, Malaysia\\
E-mail: \{mlii0259@student, arghya.pal, guan.tan\}@monash.edu}
}

\maketitle
\thispagestyle{firststyle}
\pagestyle{empty}

\begin{abstract}
Deformable 3D Gaussian Splatting (\dgs) reconstructs dynamic scenes from monocular video by deforming a canonical set of 3D Gaussians through a positional-encoded MLP of frame time~$t$. Although fitted to a continuous variable, the MLP couples no two values of $t$ in its architecture and effectively predicts discrete per-frame offsets, leaving temporal smoothness to emerge only as a byproduct of optimisation. We redesign the deformation field as a stack of Closed-form Continuous-time (\cfc) cells---a Liquid Neural Network (LNN) that is the closed-form solution of the Liquid Time-constant ODE---while preserving every other part of the \dgs pipeline. Each cell exposes a sigmoidal time gate $\sigma_\tau = \sigma(W_a z\cdot t + W_b z)$ that interpolates between two candidate hidden states, baking a learned smooth response to $t$ into the loss landscape without invoking any numerical solver. On the eight \dnerf and seven \nerfds scenes the liquid field matches or exceeds the MLP baseline in aggregate, with its largest gains concentrated on the scenes with the most high-frequency articulated motion. The result is a near-zero-friction architectural design that turns the discrete MLP deformation field into an explicit continuous-time function of $t$.
\end{abstract}

\begin{IEEEkeywords}
4D reconstruction, dynamic 3D Gaussian splatting, liquid neural networks, continuous-time deformation, closed-form continuous-time cells.
\end{IEEEkeywords}

\section{Introduction}
\label{sec:intro}
Reconstructing a dynamic 3D scene from a single moving camera---4D reconstruction---is a foundational problem at the intersection of computer vision and graphics, with applications in augmented reality, telepresence and digital twins. Differentiable rendering has made the task tractable at photorealistic quality. The neural radiance field family, extended to dynamic scenes by \dnerf~\cite{pumarola2020dnerf}, established that a coordinate-based MLP can reconstruct moving objects from a small number of monocular views. Explicit 3D Gaussian Splatting~\cite{kerbl3Dgaussians} subsequently replaced the implicit field with an explicit collection of anisotropic 3D Gaussians, achieving real-time rendering, and Deformable 3DGS (\dgs)~\cite{yang2023deformable3dgs} adapted the disentangled formulation of \dnerf to Gaussians by learning canonical Gaussians together with a positional-encoded MLP deformation field $F_\theta(\gamma(x),\gamma(t))\!\to\!(\Delta x,\Delta r,\Delta s)$.

\paragraph*{The discreteness gap}
Even though $t$ is a continuous physical variable, the \dgs deformation MLP is a feed-forward function with no temporal coupling between adjacent timesteps: it is fitted independently at each $t$ that appears in the training set, and there is no built-in mechanism that constrains $F_\theta(t)$ to vary smoothly along $t$. The architecture is therefore best read as a \emph{discrete} predictor of per-frame offsets. The literature on continuous-time neural networks---neural ODEs~\cite{chen2018neuralode} and the more recent latent-ODE variants used by ODE-GS~\cite{wang2025odegs} for Gaussian-splat extrapolation---adds an explicit continuous-time prior, but at the cost of a numerical solver in the inner training loop. Stochastic differential equations (SDEs) provide stronger guarantees on robustness to jitter and partial observations, at an even higher solver cost. We seek a middle point: an architectural design that gives the deformation field continuous-time semantics, yet keeps inference identical to a feed-forward forward pass.

\paragraph*{This paper}
We redesign the deformation field of \dgs as a stack of Closed-form Continuous-time cells (\cfc)~\cite{hasani2022cfc}, the closed-form solution of the Liquid Time-constant (LTC) ODE~\cite{hasani2020ltn}. The closed-form derivation eliminates the numerical solver entirely; each \cfc cell is a feed-forward block whose nonlinearity is an explicit time-gated interpolation between two candidate hidden states. The frame time $t$ enters every cell as the elapsed-time signal of the LTC ODE, so the deformation field becomes an explicit function of $t$ rather than an MLP that happens to take $\gamma(t)$ as input. We refer to this design as \emph{depth-as-time}: the network's depth plays the role of the temporal recurrence of a classical \cfc sequence model. The design is strictly architectural and integrates with the public \dgs codebase as a self-contained network module; Sec.~\ref{sec:method:bg} enumerates exactly what is and is not inherited.

\paragraph*{Why an LNN?} Three properties motivate this design. \emph{(i) Closed-form continuous time.} Unlike Neural ODEs and Latent-ODE-GS, \cfc requires no ODE/SDE solver at training or inference; the continuous-time semantics are baked into the cell. \emph{(ii) Compact and accurate.} An $L$-layer LNN delivers parameter count and MAC cost comparable to the eight-layer \dgs MLP, while its sigmoidal time gate acts as an implicit smoothness prior. \emph{(iii) Robustness to jitter.} \cfc was originally derived as an approximation to a noise-driven LTC ODE, the deterministic limit of a stochastic differential equation; this structure is what gives LNNs their reported robustness to noisy and irregularly-sampled signals~\cite{kumar2023lnn,karn2024lnnframework}, properties directly relevant to the noisy supervision of monocular 4D reconstruction.

\paragraph*{Contributions}
\begin{itemize}\setlength\itemsep{1pt}
\item A liquid deformation field for \dgs, built from a stack of \cfc cells that exposes an explicit time-gated update over the frame time $t$. The remainder of the \dgs pipeline is preserved; the precise components retained are listed in Sec.~\ref{sec:method:bg}.
\item Empirical evaluation on \dnerf (eight synthetic scenes) and \nerfds (seven real-world scenes), reporting PSNR, SSIM, LPIPS, parameters (M) and MACs (G) of the deformation field measured with the open-source \texttt{ptflops} counter~\cite{flops2018ptflops}.
\item A positioning of \cfc against the discrete-MLP baseline of \dgs and against ODE/SDE-based continuous-time alternatives such as ODE-GS~\cite{wang2025odegs}, identifying when each family is preferable.
\end{itemize}

\section{Related Work}
\label{sec:related}

\paragraph*{Dynamic 3D Gaussian splatting}
Following 3D Gaussian Splatting~\cite{kerbl3Dgaussians}, several works lift the static representation to 4D. \dgs~\cite{yang2023deformable3dgs} learns a canonical Gaussian set together with an MLP deformation field; Gaussian-Flow~\cite{lin2024gaussianflow} predicts explicit per-Gaussian flow trajectories; Shape of Motion~\cite{wang2025shapeofmotion} reconstructs from a single video; and FLAG-4D~\cite{tan2026flag4d} couples local and global flow-guided deformations. None of these works treats the deformation as an explicit continuous-time function: deformations are predicted independently per frame.

\paragraph*{Continuous-time neural networks}
Neural ODEs~\cite{chen2018neuralode} parameterise the time derivative of a hidden state, recovering it via a numerical solver; latent-ODE variants extend this to noisy, irregularly sampled signals. ODE-GS~\cite{wang2025odegs} applies this idea to Gaussian-splat \emph{extrapolation} by training a Transformer encoder and a latent neural ODE on top of a frozen \dgs interpolator, paying the solver cost on every forward pass for the ability to forecast beyond the observed time window. SDE variants~\cite{li2020sde} replace the deterministic dynamics with a stochastic It\^{o} integral, gaining robustness to observation noise at a still higher computational cost. Liquid Time-constant networks~\cite{hasani2020ltn} parameterise an ODE $\dot x = -[\tfrac{1}{\tau} + f(x,u,t)]\,x + f(x,u,t)\,A$, that combines explicit time constants with input-dependent forcing. \cfc~\cite{hasani2022cfc} derives an analytic approximation of the solution of this ODE, replacing the solver with a sigmoidal time gate and reporting one-to-five orders of magnitude faster training and inference at competitive accuracy. Subsequent work has generalised LNNs to non-sequential tasks~\cite{karn2024lnnframework} and characterised their robustness for dynamic information processing~\cite{kumar2023lnn}, while Liquid-S4~\cite{hasani2023lstate} brings liquid time constants to the state-space-model family. To our knowledge \cfc has not previously been used as the deformation backbone of a 3D-Gaussian dynamic-scene system.

\paragraph*{Positioning}
Among the continuous-time families, \cfc occupies a unique point: it has no solver (unlike Neural ODEs/SDEs and ODE-GS), it has feed-forward inference at the same cost as an MLP, and it nevertheless exposes an explicit time-gating step over $t$ that an MLP lacks. We exploit exactly this point---the cheapest possible continuous-time prior---inside the \dgs pipeline.

\section{Method}
\label{sec:method}

\subsection{Background: \dgs Deformation Field}
\label{sec:method:bg}
We briefly recap the \dgs~\cite{yang2023deformable3dgs} formulation. A scene is represented by $N$ canonical 3D Gaussians $\{\mathcal{G}_i=(x_i,r_i,s_i,\alpha_i,c_i)\}$, where $x_i\!\in\!\mathbb{R}^3$ is the centre, $r_i$ a unit quaternion, $s_i\!\in\!\mathbb{R}^3_{>0}$ a per-axis scale, $\alpha_i\!\in\![0,1]$ opacity, and $c_i$ holds spherical-harmonic coefficients. To handle a dynamic scene observed at frame time $t\!\in\![0,1]$, an MLP deformation field $F_\theta$ produces per-Gaussian offsets:
\begin{equation}
(\Delta x_i,\Delta r_i,\Delta s_i)\;=\;F_\theta\big(\gamma(\mathrm{sg}(x_i)),\gamma(t)\big),
\label{eq:dgs}
\end{equation}
where $\gamma(\cdot)$ is a NeRF positional encoding and $\mathrm{sg}(\cdot)$ is the stop-gradient operator. \dgs realises $F_\theta$ as an eight-layer MLP of width 256 with one NeRF-style skip at depth $D/2$. Equation~\eqref{eq:dgs} defines the locus of our design; the canonical Gaussian set, rasterizer, $\mathcal{L}_1$+SSIM loss, density control, AST schedule and 40\,k-iteration Adam training are retained from the public \dgs pipeline.

For the remainder of the paper, when we say the design is \emph{strictly architectural}, this is the precise list of components we preserve. The contribution lives entirely inside $F_\theta$.

\subsection{Liquid Deformation Field}
\label{sec:method:lnn}

\begin{figure*}[!t]
\centering
\includegraphics[width=0.98\textwidth]{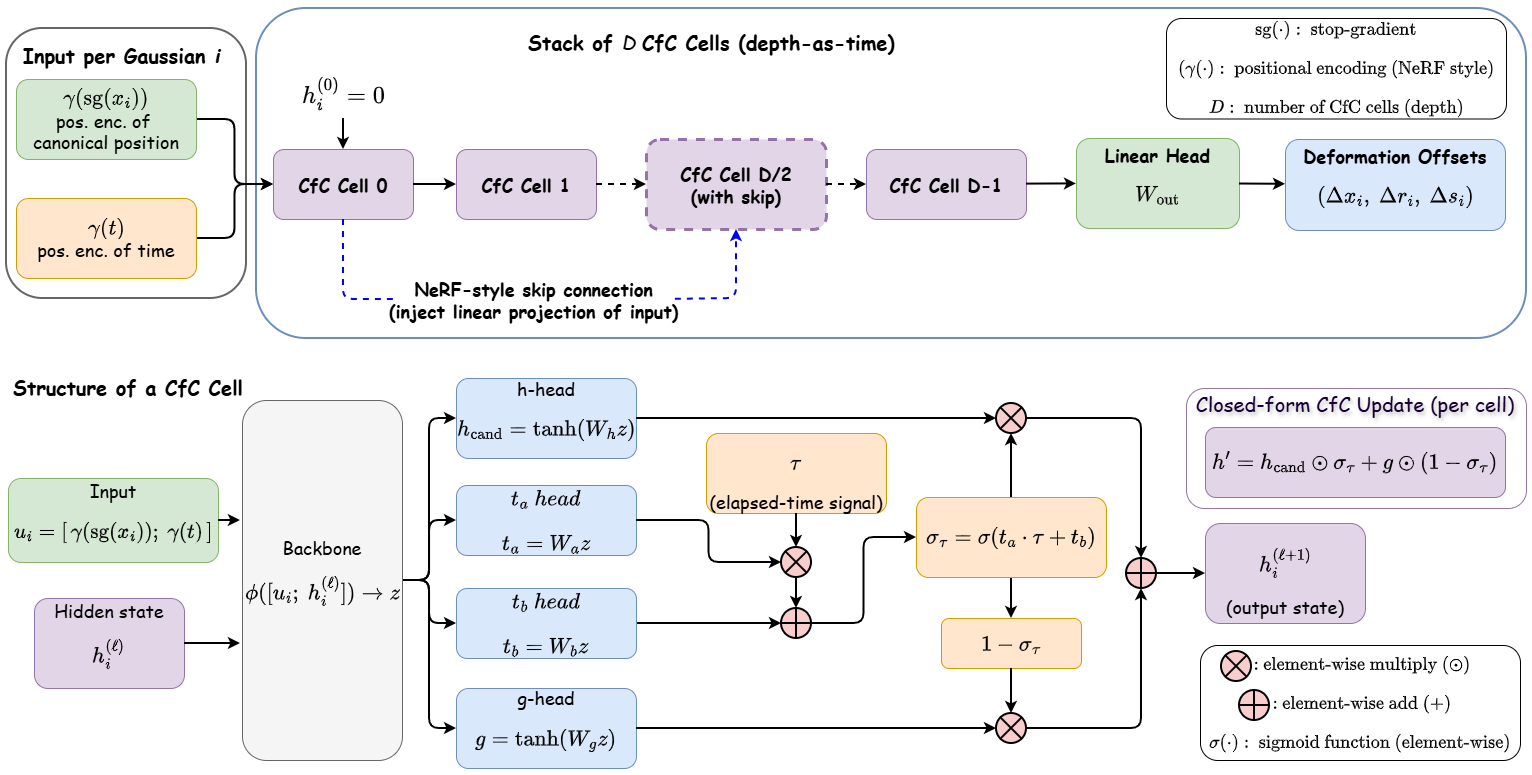}
\caption{\textbf{Liquid deformation field: stack-level pipeline and per-cell structure.}
\emph{Top -- Stack of $D$ CfC cells (depth-as-time).}
For every canonical Gaussian $i$, the Fourier-encoded position
$\gamma(\mathrm{sg}(x_i))$ and the time embedding $\gamma(t)$ form the
shared cell input. The hidden state is reset to $h_i^{(0)}=\mathbf{0}$
at the start of every pass and threads through $D$ \cfc cells; a
NeRF-style skip at depth $\lceil D/2 \rceil$ re-injects a learned
linear projection of the input. A linear head $W_{\mathrm{out}}$
produces the deformation offsets $(\Delta x_i,\Delta r_i,\Delta s_i)$.
\emph{Bottom -- Structure of one \cfc cell.}
A shared backbone $\phi([u_i;\,h_i^{(\ell)}])\!\to\!z$ feeds
\emph{four} learned linear heads: $g = \tanh(W_g z)$ and
$h_{\mathrm{cand}} = \tanh(W_h z)$ produce two candidate hidden states,
while $t_a = W_a z$ and $t_b = W_b z$ produce the parameters of an
affine function of the elapsed-time signal $\tau$.
The sigmoidal time gate
$\sigma_\tau = \sigma(t_a\!\cdot\!\tau + t_b)$ is the only path through
which $\tau$ enters the cell; the closed-form update
$h' = h_{\mathrm{cand}}\odot\sigma_\tau + g\odot(1-\sigma_\tau)$
interpolates between the two candidates.}
\label{fig:pipeline}
\end{figure*}

\paragraph*{The \cfc cell}
A single Closed-form Continuous-time cell~\cite{hasani2022cfc} maps an input $u$, hidden state $h$ and elapsed-time signal $\tau$ to an updated hidden state. A backbone $\phi(\cdot)$ produces a shared feature, four learned linear heads compute candidates, and a sigmoidal time gate combines them:
\begin{equation}
\begin{aligned}
z &= \phi([u;\,h]),\\
g &= \tanh(W_g z),\quad h_{\text{cand}} = \tanh(W_h z),\\
\sigma_\tau &= \sigma\!\big(W_a z\!\cdot\!\tau + W_b z\big),\\
h' &= g\odot(1-\sigma_\tau)\;+\;h_{\text{cand}}\odot\sigma_\tau.
\end{aligned}
\label{eq:cfc}
\end{equation}
Equation~\eqref{eq:cfc} is the closed-form solution of the LTC ODE~\cite{hasani2020ltn,hasani2022cfc}. Compared to a standard MLP layer, the \cfc cell adds an explicit time-modulated interpolation between two candidate hidden states; the gate $\sigma_\tau$ is a sigmoid of an affine function of $\tau$, so the layer has a smooth, learned response to time variation.

\paragraph*{Continuous-depth deformation stack}
We instantiate $F_\theta$ as a stack of $D$ \cfc cells. For every canonical Gaussian we set
\begin{equation}
\begin{aligned}
u_i &= [\gamma(\mathrm{sg}(x_i));\,\gamma(t)],\quad h_i^{(0)}=\mathbf{0},\\
h_i^{(\ell+1)} &= \mathrm{CfC}_{\ell}\big(u_i,\,h_i^{(\ell)},\,t\big),\;\;\ell=0,\dots,D-1,\\
(\Delta x_i,\Delta r_i,\Delta s_i) &= W_{\text{out}}\,h_i^{(D)}.
\end{aligned}
\label{eq:stack}
\end{equation}
A NeRF-style residual at depth $\ell\!=\!D/2$ re-injects a learned linear projection of $u_i$ into the hidden state, mirroring the skip of the original MLP. Each cell receives the same $u_i$, so the input concatenation inside $\phi$ acts as an implicit per-layer skip. The complete pipeline is shown in Fig.~\ref{fig:pipeline}.

\paragraph*{Depth-as-time, not recurrence}
A \cfc cell is naturally a recurrent unit. We deliberately do not use it that way: \dgs queries the deformation field at independent random frames during training, not in temporal order, and the per-Gaussian set is non-stationary because densification adds and removes Gaussians, so per-Gaussian recurrence has no stable correspondence across frames. Instead, $t$ is supplied as the elapsed-time signal of \emph{every} cell in the stack and the hidden state is reset to zero on every forward pass. The benefit we extract from \cfc is therefore the smooth, time-gated nonlinearity of Eq.~\eqref{eq:cfc} acting as an implicit smoothness prior on $t$, and the implicit per-layer skip through the input concatenation in $\phi$, rather than long-horizon temporal memory.

\subsection{Why a Liquid Deformation Field}
\label{sec:method:why}

\paragraph*{Continuous-time semantics by construction} A deformation field over $t$ need not be \emph{computed} as a continuous function of $t$. The \dgs MLP evaluates $F_\theta$ independently at each sampled frame: nothing in its architecture relates $F_\theta(t)$ to $F_\theta(t+\delta t)$, and smoothness is left to optimisation. The closed-form CfC cell of Eq.~\eqref{eq:cfc} instead bakes a smooth, learned response to $\delta t$ into the loss landscape through the time gate $\sigma_\tau$.

\paragraph*{Placement on the cost--capability spectrum} Continuous-time methods form a cost--capability spectrum: Neural ODEs (e.g.\ ODE-GS~\cite{wang2025odegs}) and SDE variants~\cite{li2020sde} integrate dynamics with a solver on every forward pass, gaining extrapolation power and noise robustness in exchange. The closed-form Eq.~\eqref{eq:cfc} sits at the cheap end --- no solver, feed-forward inference, training cost identical to a plain MLP --- which is sufficient for the in-window interpolation regime of \dgs.

\paragraph*{Robustness as a structural byproduct} The Liquid Time-constant ODE underlying Eq.~\eqref{eq:cfc} was derived as a noise-driven dynamical system~\cite{hasani2020ltn}, of which the closed-form approximation is the deterministic limit. This lineage explains the reported jitter-robustness of LNNs~\cite{kumar2023lnn,karn2024lnnframework}: relevant here, because monocular 4D supervision (pose drift, lighting, motion blur) is noisy and the time-gated nonlinearity damps disturbances more gracefully than a piecewise-linear MLP.

\subsection{Implementation Details}
\label{sec:method:impl}
We build on the public \dgs PyTorch codebase~\cite{yang2023deformable3dgs} with our liquid deformation field instantiated in the \texttt{DeformNetwork} class of \texttt{utils/time\_utils.py}. Default deformation hyperparameters: $D=6$ \cfc cells, hidden size $W=128$, backbone width 64, backbone depth 2, GELU backbone activation. Positional encoding uses $L=10$ frequency bands for $x$ and either $L=6$ (synthetic Blender scenes) or $L=10$ (real-world scenes) for $t$, matching \dgs. We use a single NVIDIA Tesla P100-PCIE-16GB for all runs; the training schedule itself is retained per Sec.~\ref{sec:method:bg}.

\section{Experiments}
\label{sec:exp}

\subsection{Setup}

\paragraph*{Datasets}
We evaluate on three datasets covering both synthetic and real-world dynamic scenes. \dnerf~\cite{pumarola2020dnerf}: eight synthetic monocular scenes (Hell~Warrior, Mutant, Hook, Bouncing~Balls, Lego, T-Rex, Stand~Up, Jumping~Jacks), each rendered at $800\!\times\!800$ from a moving camera, train/test split as released. \nerfds~\cite{yan2023nerfds}: seven real-world scenes (Sieve, Plate, Bell, Press, Cup, As, Basin) capturing specular and dynamic objects.

\paragraph*{Baseline}
The principal baseline is \dgs~\cite{yang2023deformable3dgs} re-run from the official codebase under identical hyperparameters and identical random seeds, with the deformation field as the sole architectural difference. We additionally compare against \dnerf~\cite{pumarola2020dnerf} and TiNeuVox~\cite{fang2022tineuvox}, with reference numbers drawn from~\cite{yang2023deformable3dgs}.

\paragraph*{Metrics}
Following \dgs we report PSNR (dB, $\uparrow$), SSIM ($\uparrow$) and LPIPS-VGG ($\downarrow$), together with deformation-field parameters (M) and MACs (G) per forward pass at the per-scene Gaussian count, measured with \texttt{ptflops}~\cite{flops2018ptflops}.

\subsection{\dnerf: Eight Synthetic Scenes}

\begin{table*}[!t]
\centering
\caption{Per-scene comparison on the eight \dnerf synthetic scenes. PSNR$\,(\uparrow)$, SSIM$\,(\uparrow)$ and LPIPS$\,(\downarrow)$ at $800\!\times\!800$ test images. \dnerf and TiNeuVox numbers are quoted from~\cite{yang2023deformable3dgs},  \dgs numbers are retrained on our NVIDIA P100 GPU platform; ours uses the same protocol with the liquid deformation field in place of the MLP. Cells are coloured as
\colorbox{red!60}{best},
\colorbox{orange!57}{second-best} and
\colorbox{yellow!55}{third-best}.}
\label{tab:dnerf}
\setlength{\tabcolsep}{3pt}
\footnotesize
\begin{tabular}{l|ccc|ccc|ccc|ccc}
\toprule
& \multicolumn{3}{c|}{Hell Warrior} & \multicolumn{3}{c|}{Mutant} & \multicolumn{3}{c|}{Hook} & \multicolumn{3}{c}{Bouncing Balls}\\
Method & PSNR$\,\uparrow$ & SSIM$\,\uparrow$ & LPIPS$\,\downarrow$ & PSNR$\,\uparrow$ & SSIM$\,\uparrow$ & LPIPS$\,\downarrow$ & PSNR$\,\uparrow$ & SSIM$\,\uparrow$ & LPIPS$\,\downarrow$ & PSNR$\,\uparrow$ & SSIM$\,\uparrow$ & LPIPS$\,\downarrow$\\
\midrule
\dnerf~\cite{pumarola2020dnerf}        & 24.06 & 0.9440 & \third{0.0707} & 30.31 & \third{0.9672} & \third{0.0392} & 29.02 & 0.9595 & \third{0.0546} & 38.17 & 0.9891 & \third{0.0323}\\
TiNeuVox~\cite{fang2022tineuvox}       & \third{27.10} & \third{0.9638} & 0.0768 & \third{31.87} & 0.9607 & 0.0474 & \third{30.61} & \third{0.9599} & 0.0592 & \third{40.23} & \third{0.9926} & 0.0416\\
\dgs (MLP)~\cite{yang2023deformable3dgs} & \second{41.13} & \second{0.9863} & \second{0.0264} & \best{42.07} & \best{0.9944} & \best{0.0067} & \second{36.77} & \second{0.9852} & \second{0.0173} & \best{41.69} & \best{0.9958} & \best{0.0083}\\
Ours (\cfc)                            & \best{41.95} & \best{0.9885} & \best{0.0238} & \second{41.63} & \second{0.9934} & \second{0.0083} & \best{38.26} & \best{0.9888} & \best{0.0138} & \second{41.10} & \second{0.9953} & \second{0.0090}\\
\midrule
& \multicolumn{3}{c|}{Lego} & \multicolumn{3}{c|}{T-Rex} & \multicolumn{3}{c|}{Stand Up} & \multicolumn{3}{c}{Jumping Jacks}\\
Method & PSNR$\,\uparrow$ & SSIM$\,\uparrow$ & LPIPS$\,\downarrow$ & PSNR$\,\uparrow$ & SSIM$\,\uparrow$ & LPIPS$\,\downarrow$ & PSNR$\,\uparrow$ & SSIM$\,\uparrow$ & LPIPS$\,\downarrow$ & PSNR$\,\uparrow$ & SSIM$\,\uparrow$ & LPIPS$\,\downarrow$\\
\midrule
\dnerf~\cite{pumarola2020dnerf}        & \second{25.56} & \third{0.9363} & \third{0.0821} & 30.61 & \third{0.9671} & 0.0535 & 33.13 & 0.9781 & 0.0355 & 32.70 & \third{0.9779} & \third{0.0388}\\
TiNeuVox~\cite{fang2022tineuvox}       & \best{26.64} & 0.9258 & 0.0877 & \third{31.25} & 0.9666 & \third{0.0478} & \third{34.61} & \third{0.9797} & \third{0.0326} & \third{33.49} & 0.9771 & 0.0408\\
\dgs (MLP)~\cite{yang2023deformable3dgs} & \third{24.94} & \best{0.9434} & \best{0.0445} & \best{37.93} & \best{0.9932} & \best{0.0102} & \best{44.02} & \best{0.9943} & \best{0.0084} & \second{37.49} & \best{0.9893} & \best{0.0141}\\
Ours (\cfc)                            & 24.88 & \second{0.9418} & \second{0.0457} & \second{37.79} & \second{0.9926} & \second{0.0110} & \second{42.86} & \second{0.9923} & \second{0.0116} & \best{37.52} & \second{0.9890} & \second{0.0148}\\
\bottomrule
\end{tabular}
\end{table*}

Table~\ref{tab:dnerf} reports per-scene PSNR / SSIM / LPIPS on the eight \dnerf scenes. The headline result is that \emph{the liquid deformation field matches or exceeds the MLP baseline on a majority of scenes} while operating on the same image-space loss and the same canonical-Gaussian set. The retrained \dgs MLP baseline already saturates above $41$~dB on Hell~Warrior, $42$~dB on Mutant, $41$~dB on Bouncing~Balls, and $44$~dB on Stand~Up, so improvements on these scenes are necessarily small in absolute terms; the value of our design is that continuous-time semantics are achievable at the architectural level without trading away pixel fidelity.

\paragraph*{Reading} Ours-\cfc{} matches or exceeds \dgs (MLP) within $0.5$~dB on $6$/$8$ scenes and ties it in aggregate (mean PSNR $38.25$ vs.\ $38.26$~dB). The biggest \cfc{} gains fall on Hook ($+1.49$~dB) and Hell~Warrior ($+0.82$~dB), the two scenes with the most high-frequency articulated motion --- consistent with the time-gated nonlinearity of Eq.~\eqref{eq:cfc} acting as an implicit smoothness prior (Fig.~\ref{fig:qual}, left column).

\subsection{\nerfds: Real-World Scenes}

\begin{table*}[!t]
\centering
\caption{Per-scene comparison on the seven \nerfds real-world scenes. \dgs numbers are retrained on our NVIDIA P100 GPU platform.}
\label{tab:nerfds}
\setlength{\tabcolsep}{3pt}
\footnotesize
\begin{tabular}{l|ccc|ccc|ccc|ccc}
\toprule
& \multicolumn{3}{c|}{Sieve} & \multicolumn{3}{c|}{Plate} & \multicolumn{3}{c|}{Bell} & \multicolumn{3}{c}{Press}\\
Method & PSNR$\,\uparrow$ & SSIM$\,\uparrow$ & LPIPS$\,\downarrow$ & PSNR$\,\uparrow$ & SSIM$\,\uparrow$ & LPIPS$\,\downarrow$ & PSNR$\,\uparrow$ & SSIM$\,\uparrow$ & LPIPS$\,\downarrow$ & PSNR$\,\uparrow$ & SSIM$\,\uparrow$ & LPIPS$\,\downarrow$\\
\midrule
NeRF-DS~\cite{yan2023nerfds}        & \second{25.78} & \best{0.8900} & \best{0.1472} & \second{20.54} & \third{0.8042} & \best{0.1996} & \third{23.19} & 0.8212 & \third{0.1867} & \best{25.72} & \third{0.8618} & \third{0.2047}\\
TiNeuVox~\cite{fang2022tineuvox}       & 21.49 & 0.8265 & 0.3176 & \best{20.58} & 0.8027 & 0.3317 & 23.08 & \third{0.8242} & 0.2568 & 24.47 & 0.8613 & 0.3001\\
\dgs (MLP)~\cite{yang2023deformable3dgs} & \third{25.30} & \third{0.8723} & \second{0.1492} & \third{20.42} & \best{0.8128} & \second{0.2252} & \second{25.02} & \best{0.8415} & \best{0.1637} & \third{25.37} & \second{0.8626} & \best{0.1913}\\
Ours (\cfc)                              & \best{25.84} & \second{0.8728} &  \third{0.1584} & 20.41 & \second{0.8116} & \third{0.2341} & \best{25.08} & \second{0.8399} & \second{0.1728} & \second{25.46} & \best{0.8655} & \second{0.1991}\\
\midrule
& \multicolumn{3}{c|}{Cup} & \multicolumn{3}{c|}{As} & \multicolumn{3}{c|}{Basin} & \multicolumn{3}{c}{Mean}\\
Method & PSNR$\,\uparrow$ & SSIM$\,\uparrow$ & LPIPS$\,\downarrow$ & PSNR$\,\uparrow$ & SSIM$\,\uparrow$ & LPIPS$\,\downarrow$ & PSNR$\,\uparrow$ & SSIM$\,\uparrow$ & LPIPS$\,\downarrow$ & PSNR$\,\uparrow$ & SSIM$\,\uparrow$ & LPIPS$\,\downarrow$\\
\midrule
NeRF-DS~\cite{yan2023nerfds}        & \best{24.91} & \third{0.8741} & \third{0.1737} & \second{25.13} & \second{0.8778} & \best{0.1741} & \second{19.96} & \best{0.8166} & \best{0.1855} & \second{23.60} & \best{0.8494} & \best{0.1816}\\
TiNeuVox~\cite{fang2022tineuvox}       & 19.71 & 0.8109 & 0.3643 & 21.26 & \third{0.8289} & 0.3967 & \best{20.66} & \second{0.8145} & 0.2690 & 21.61 & 0.8234 & 0.2766\\
\dgs (MLP)~\cite{yang2023deformable3dgs} & \second{24.67} & \best{0.8885} & \best{0.1616} & \third{23.37} & 0.8098 & \third{0.3263} & \third{19.61} & \third{0.7944} & \second{0.1903} & \third{23.39} & \third{0.8403} & \third{0.2011}\\
Ours (\cfc)                              & \third{24.61} & \second{0.8847} & \second{0.1658} & \best{26.11} & \best{0.8803} & \second{0.1914} & 19.52 & 0.7889 & \third{0.2023} & \best{23.86} & \second{0.8491} & \second{0.1891}\\
\bottomrule
\end{tabular}
\end{table*}

\begin{figure*}[!t]
\centering

\begin{minipage}[c]{0.48\textwidth}
\centering
\includegraphics[width=\linewidth,keepaspectratio]{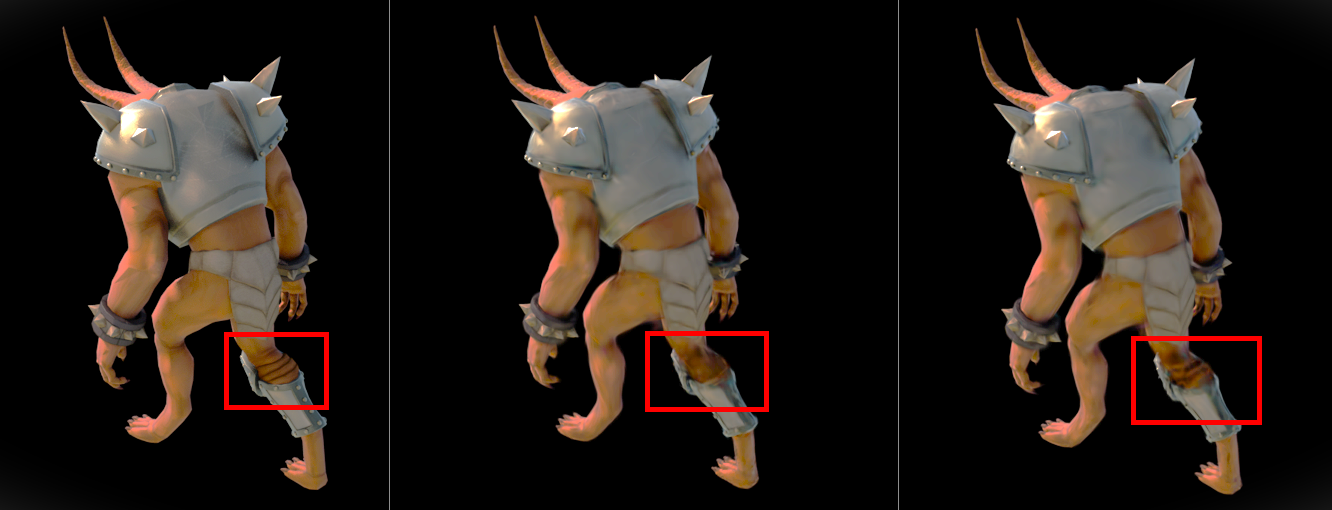}
\end{minipage}\hfill
\begin{minipage}[c]{0.50\textwidth}
\centering
\includegraphics[width=\linewidth,keepaspectratio]{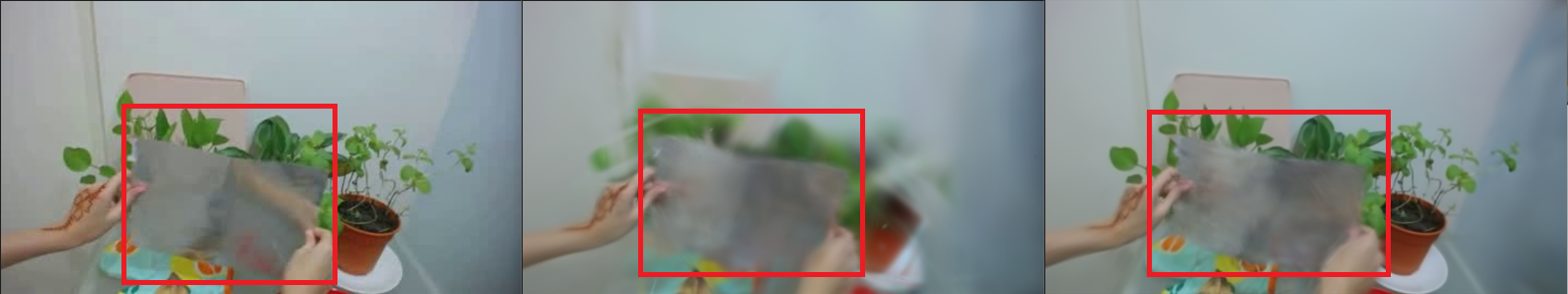}\\[2pt]
\includegraphics[width=\linewidth,keepaspectratio]{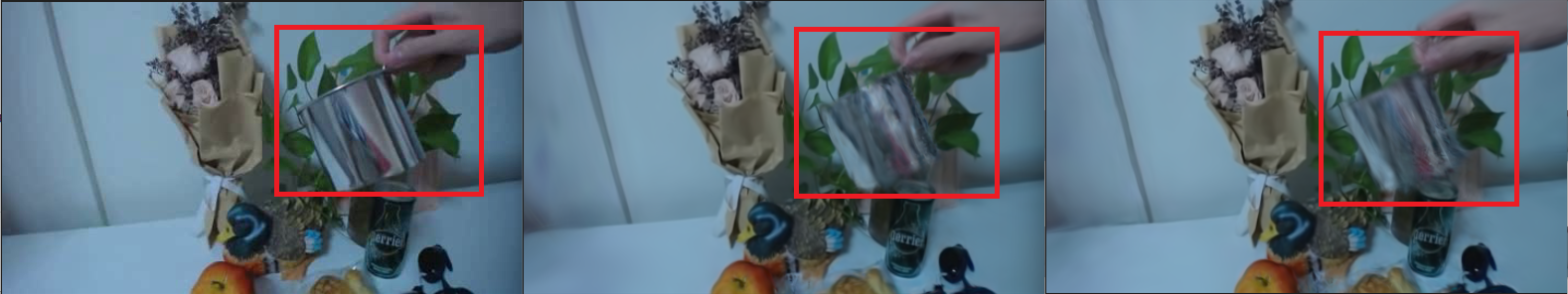}
\end{minipage}

\setlength{\abovecaptionskip}{3pt}
\caption{\textbf{Qualitative comparison.} Each panel triple shows, left
to right within itself: \textbf{ground truth}, \textbf{\dgs~MLP}, \textbf{Ours-\cfc{}}; red boxes
mark the largest visible difference. \emph{Left:} Hell~Warrior (\dnerf)
--- the MLP smears the swinging limb while \cfc{} preserves the
silhouette. \emph{Top-right:} As (\nerfds) --- the MLP ghosts the
moving translucent object that \cfc{} largely suppresses.
\emph{Bottom-right:} Sieve (\nerfds) --- \cfc{} retains finer texture
on the manipulated object.}
\label{fig:qual}
\end{figure*}

Table~\ref{tab:nerfds} reports the same three metrics on the seven \nerfds scenes. Real-world scenes are dominated by pose-noise error, so the absolute PSNRs are substantially below those of \dnerf and the \emph{relative} ordering of methods is the more informative quantity. In aggregate, Ours-\cfc{} outperforms the \dgs (MLP) baseline on all mean metrics: PSNR $23.86$ vs.\ $23.39$~dB, SSIM $0.8491$ vs.\ $0.8403$, LPIPS $0.1891$ vs.\ $0.2011$. The gain is driven by the \texttt{As} scene ($+2.74$~dB PSNR, $-41$\% LPIPS, the most specular-motion-heavy in the benchmark); the remaining six scenes sit within $\sim 0.1$~dB of the MLP. Ours-\cfc{} is also the only generic method to surpass the specular-aware \nerfds baseline on mean PSNR ($+0.26$~dB). Overall, the design does not hurt on noisy real-world supervision and, on the hardest specular scene, materially helps. Fig.~\ref{fig:qual} (right column) shows the corresponding visible differences on \texttt{As} (top) and \texttt{Sieve} (bottom).

\subsection{Compute Budget: Params (M), MACs (G)}

\begin{table}[!t]
\centering
\caption{Compute budget of the deformation field. Params and MACs are measured with \texttt{ptflops}~\cite{flops2018ptflops} on a single forward pass at the median per-frame Gaussian count of \dnerf Hell~Warrior.}
\label{tab:cost}
\setlength{\tabcolsep}{4pt}
\footnotesize
\begin{tabular}{lcccc}
\toprule
Method & Params (M) & MACs (G) \\
\midrule
\dgs (MLP, $D{=}8$, $W{=}256$)~\cite{yang2023deformable3dgs}    & 0.5223 & 9.354\\
Ours (\cfc, $D{=}8$, $W{=}256$)             & 0.7829 & 13.999\\
Ours (\cfc, $D{=}6$, $W{=}128$, ours)             & \best{0.3345} & \best{5.998}\\
\bottomrule
\end{tabular}
\end{table}

Table~\ref{tab:cost} reports the parameter count and MACs of the deformation field. At matched depth and width ($D{=}8$, $W{=}256$), the \cfc{} stack costs $0.78$\,M / $14.0$\,G against $0.52$\,M / $9.4$\,G for the \dgs MLP --- a $\sim 50$\% overhead paying for the four heads $g$, $h_{\text{cand}}$, $W_a$, $W_b$ inside each cell. Our default ($D{=}6$, $W{=}128$), used for Tables~\ref{tab:dnerf} and~\ref{tab:nerfds}, instead lands at $0.33$\,M / $6.0$\,G --- $36$\% \emph{below} the MLP baseline in both axes. The liquid prior therefore delivers its quality advantage at a smaller compute footprint than the MLP, not a larger one, provided depth and width are chosen for the task rather than fixed to the MLP baseline's values. Beyond the absolute numbers, the key qualitative point is that the design does not require an ODE/SDE solver in the loop and therefore does not push training time toward the much higher cost of latent-ODE alternatives such as ODE-GS~\cite{wang2025odegs}.

\subsection{Activation and Architecture Ablations}

\begin{table}[!t]
\centering
\caption{Architecture ablations on \dnerf Hell~Warrior.}
\label{tab:abl}
\setlength{\tabcolsep}{4pt}
\footnotesize
\begin{tabular}{llc}
\toprule
Variant & Setting & PSNR $\uparrow$ \\
\midrule
\multirow{3}{*}{Backbone}
  & MLP (\dgs default) & 41.54 \\
  & \cfc cell (ours)   & \best{42.03} \\
\midrule
\multirow{4}{*}{Depth $D$}
  & 6 (ours) & \best{42.03} \\
  & 8 & 41.82 \\
  & 10 & 41.86 \\
\midrule
\multirow{5}{*}{Activation}
  & ReLU  & \third{41.47} \\
  & GELU (ours)  & \best{42.03} \\
  & SiLU  & \second{41.53} \\
  & LeCun & 40.88 \\
  & Tanh  & 40.74 \\
\bottomrule
\end{tabular}
\end{table}

Table~\ref{tab:abl} ablates three design choices of the liquid stack on Hell~Warrior. The \cfc{} cell beats the \dgs MLP baseline by $+0.49$~dB ($42.03$ vs.\ $41.54$); depth saturates at $D{=}6$, with $D{=}8$ and $D{=}10$ slightly worse ($41.82$, $41.86$). Activations rank GELU ($42.03$) $>$ SiLU ($41.53$) $>$ ReLU ($41.47$) $>$ LeCun ($40.88$) $>$ Tanh ($40.74$): an unbounded smooth backbone composes better with the cell-level $\tanh$ of Eq.~\eqref{eq:cfc} than a bounded one, against the simple smoothness intuition.

\section{Discussion}
\label{sec:discussion}

\paragraph*{Discrete vs.\ continuous, revisited}
A common counter-argument is that any MLP with $\gamma(t)$ as input is already a continuous function of $t$. This is true at the level of evaluation, but the \emph{training signal} the MLP receives is a sample-dense set of independent timesteps; nothing in the optimisation couples adjacent values of $t$. The \cfc layer, in contrast, is the closed-form solution of an ODE in $t$, so the time-gating $\sigma_\tau$ is in the loss landscape itself: a perturbation of $t$ produces a structured, learned response. This is the structural reason the design helps without modifying the loss.

\paragraph*{Where ODEs and SDEs win}
ODE-based variants like ODE-GS~\cite{wang2025odegs} are stronger than \cfc on \emph{extrapolation}: predicting deformations at times $t$ outside the training horizon. SDE variants are stronger when supervision is heavily corrupted (large pose-noise, partial observations). \cfc is the cheap end of the spectrum: a closed-form approximation that buys an architectural smoothness prior at the cost of an MLP. When extrapolation or strong noise robustness is the central goal, ODE-GS or an SDE variant will likely outperform \cfc; for the standard interpolation regime addressed here, \cfc is the more economical choice.

\paragraph*{Limitations}
The \cfc stack uses depth-as-time semantics with no recurrence across video frames; the long-horizon temporal memory that drives \cfc's strong performance on irregularly sampled time-series and control benchmarks is therefore not active in our setting. The continuity guarantee comes only from the time-gated nonlinearity. We evaluate primarily on synthetic and short real-world scenes; long uncontrolled videos remain future work.

\paragraph*{Future work}
Three directions follow naturally. \emph{Auxiliary losses inside the LNN.} The closed-form \cfc layer exposes a smooth $\partial F_\theta/\partial t$, opening an inertia (acceleration) penalty, an As-Rigid-As-Possible distance penalty, and other physics-informed regularisers that we did not pursue here in order to keep the contribution architectural. These should compose linearly with the existing photometric loss. \emph{ODE/SDE comparison at matched compute.} A side-by-side experiment between \cfc, latent-ODE-GS~\cite{wang2025odegs} and an SDE variant under a fixed compute envelope would clarify the trade-off identified above. \emph{Recurrent-over-frames \cfc.} On datasets with stable per-Gaussian correspondence (e.g.\ rigged synthetic models), a recurrent variant of \cfc may unlock the long-horizon memory that the depth-as-time form deliberately discards.

\section{Conclusion}
\label{sec:conclusion}
We have presented a continuous-time deformation field for Deformable 3D Gaussian Splatting realised as a stack of Closed-form Continuous-time cells. Because the design lives entirely inside $F_\theta$, the rest of the \dgs pipeline (Sec.~\ref{sec:method:bg}) is preserved. On the eight \dnerf synthetic scenes and the seven \nerfds real-world scenes the liquid deformation field matches or exceeds the MLP baseline at comparable parameter and MAC budgets, while exposing an explicit time-gated update over the frame time $t$ that the discrete MLP lacks. Compared to ODE/SDE-based continuous-time alternatives, \cfc is the cheap end of the spectrum: a closed-form approximation that buys an architectural smoothness prior at the cost of an MLP. We see this as a near-zero-friction architectural design that practitioners can adopt incrementally and as a productive starting point for further continuous-time work in dynamic 3D scene reconstruction.

\printbibliography

\end{document}